\documentclass[sigconf]{acmart}
\usepackage{multirow}
\usepackage{makecell}
\usepackage{adjustbox} 
\usepackage{graphicx}

\copyrightyear{2023}
\acmYear{2023}
\setcopyright{acmlicensed}\acmConference[ICMI '23]{INTERNATIONAL CONFERENCE ON MULTIMODAL INTERACTION}{October 9--13, 2023}{Paris, France}
\acmBooktitle{INTERNATIONAL CONFERENCE ON MULTIMODAL INTERACTION (ICMI '23), October 9--13, 2023, Paris, France}
\acmPrice{15.00}
\acmDOI{10.1145/3577190.3614117}
\acmISBN{979-8-4007-0055-2/23/10}

\acmSubmissionID{1335}  

\begin{document}

\title{MMASD: A Multimodal Dataset for Autism Intervention Analysis}

\author{Jicheng Li}
\orcid{0000-0003-2564-6337}
\affiliation{%
  \institution{University of Delaware}
  \streetaddress{18 Amstel Ave}
  \city{Newark}
  \state{DE}
  \country{United States}
  \postcode{19716}
}
\email{lijichen@udel.edu}

\author{Vuthea Chheang}
\affiliation{%
  \institution{University of Delaware}
  \city{Newark}
  \state{DE}
  \country{United States}
}
\email{vuthea@udel.edu}
\orcid{0000-0001-5999-4968}

\author{Pinar Kullu}
\affiliation{%
  \institution{University of Delaware}
  \city{Newark}
  \state{DE}
  \country{United States}
}
\email{pkullu@udel.edu}
\orcid{0000-0003-3396-8782}

\author{Eli Brignac}
\affiliation{%
  \institution{University of Delaware}
  \city{Newark}
  \state{DE}
  \country{United States}
}
\email{ebrignac@udel.edu}
\orcid{0009-0000-3922-4376}

\author{Zhang Guo}
\affiliation{%
  \institution{University of Delaware}
  \city{Newark}
  \state{DE}
  \country{United States}
}
\email{guozhang@udel.edu}
\orcid{0000-0003-1599-571X}

\author{Kenneth E. Barner}
\affiliation{%
  \institution{University of Delaware}
  \city{Newark}
  \state{DE}
  \country{United States}
}
\email{barner@udel.edu}
\orcid{0000-0002-0936-7840}

\author{Anjana Bhat}
\affiliation{%
  \institution{University of Delaware}
  \streetaddress{540 S. College Ave} 
  \city{Newark}
  \state{DE}
  \country{United States}
  \postcode{19713}
}
\email{abhat@udel.edu}
\orcid{0000-0002-0771-0967}
\author{Roghayeh Leila Barmaki}
\authornote{Correspondence: Roghayeh Leila Barmaki (rlb@udel.edu)}
\affiliation{%
  \institution{University of Delaware}
  \city{Newark}
  \state{DE}
  \country{United States}
}
\email{rlb@udel.edu}
\orcid{0000-0002-7570-5270}

\renewcommand{\shortauthors}{Li \emph{et al.}}
\newcommand{\etal}{\textit{et al. }}

\begin{abstract}
 Autism spectrum disorder (ASD) is a developmental disorder characterized by significant impairments in social communication and difficulties perceiving and presenting communication signals. 
Machine learning techniques have been widely used to facilitate autism studies and assessments. 
However, computational models are primarily
concentrated on very specific analysis and validated on private, non-public datasets in the autism community, which limits comparisons across models due to privacy-preserving data-sharing complications.
This work presents a novel open source privacy-preserving dataset, \textbf{MMASD} as a \textbf{M}ulti\textbf{M}odal \textbf{ASD} benchmark dataset, collected from play therapy interventions for children with autism. 
The MMASD includes data from \textbf{32} children with ASD, and \textbf{1,315} data samples segmented from more than \textbf{100} hours of intervention recordings. 
To promote the privacy of children while offering public access, each sample consists of four privacy-preserving modalities, some of which are derived from original videos: \textbf{(1)} optical flow, \textbf{(2)} 2D skeleton, \textbf{(3)} 3D skeleton, and \textbf{(4)} clinician ASD evaluation scores of children. 
MMASD aims to assist researchers and therapists in understanding children's cognitive status, monitoring their progress during therapy, and customizing the treatment plan accordingly. 
It also inspires downstream social tasks such as action quality assessment and interpersonal synchrony estimation. 
The dataset is publicly accessible via the \href{https://sites.udel.edu/hci-lab/mmasd-project/}{MMASD project website}. 

\end{abstract}




\begin{teaserfigure}
\includegraphics[width=\textwidth]{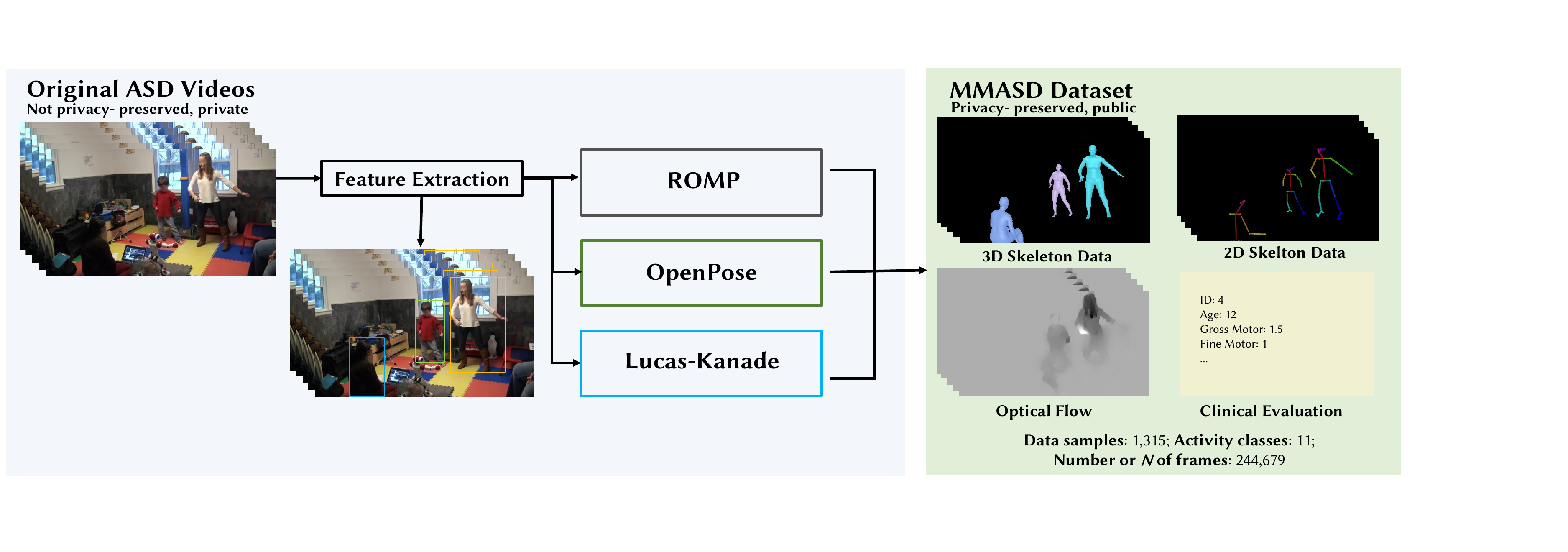}
\caption{MMASD provides multiple multimodal privacy-preserving features derived from original videos via ROMP \cite{ROMP}, OpenPose \cite{cao2019openpose}, and Lucas-Kanade \cite{lucas1981iterative}, including optical flow, 2D / 3D skeleton, and clinical evaluation results. The dataset is publicly accessible for addressing research questions centered on social and behavioral interactions of children with ASD in playful group activities.}
\Description{taser}
\label{fig:teaser}
\end{teaserfigure}


\maketitle

\section{Introduction}
Autism spectrum disorder (ASD) is a neurodevelopmental disorder characterized by significant impairments in social communication, as well as difficulties in perceiving and expressing communication cues.
Approximately 1 in 54 children are on the spectrum in the United States, resulting in over 1 million affected individuals nationwide \cite{baio2018prevalence}. 
Primary treatment of ASD includes behavioral and psychosocial interventions accompanied by prescribed medications. 
Behavioral and psychosocial interventions facilitate social and communication development, while medication helps control associated symptoms and comorbid problems \cite{billing2020dream}. 
Specifically, psychosocial interventions are diverse in content and can vary in curriculum, structure, discipline, and theme. 
Prevailing therapeutic interventions include applied behavior analysis and robot-assisted therapy, both of which can provide valuable data for analyzing children's mental development and developing individualized treatment plans.

Various studies have widely used machine learning techniques to facilitate autism research \cite{li2022dyadic, li2022pose, guo2023social, marinoiu20183d, Pandey2020Guided, Chen2019AttentionBasd, de2020computer}. 
Compared to traditional methods that rely heavily on human expertise and experience, machine learning approaches can help reduce the need for human labor and associated costs while still achieving decent performance.
The autism community has benefited from machine learning techniques in many areas, including but not limited to autism diagnosis \cite{Chen2019AttentionBasd, wall2012use}, emotion recognition \cite{li21twostage, marinoiu20183d}, and movement pattern assessment \cite{li2022dyadic, li2022pose, Li2021improving, Pandey2020Guided}. 

In machine learning research, it is widely accepted to follow a research pipeline that involves developing, applying, and comparing models across multiple benchmark datasets to ensure fair comparisons in performance. 
However, in the autism community, commonly recognized benchmarks, especially for behavior analysis and activity understanding, are limited due to privacy concerns. 
Typically, models are validated on a private dataset, and data-sharing roadblocks can restrict the comparison between models.
In this sense, the availability of publicly accessible datasets is a crucial first step for the autism community since it allows cutting-edge machine-learning techniques to be trained and validated on ASD datasets.
Although some studies have already been conducted, there is still a significant lack of \textit{publicly available, multimodal} datasets that can be used to analyze the \textit{full-body movements} of children during therapeutic interventions.

To overcome some of these ASD data-sharing challenges,  we propose a publicly available multimodal ASD dataset, \textbf{MMASD}\footnote{The MMASD Dataset is accessible via \url{
https://sites.udel.edu/hci-lab/mmasd-project/
}.}. 
MMASD maintains privacy while retaining essential movement features by providing optical flow, 2D and 3D skeletons that are derived from the original play therapy videos, thereby avoiding the exposure of sensitive and identifiable raw video footage.
Additionally, it includes clinician evaluation results of each child, such as motor function scores.
We also provide the intervention activity class labels for overall scene understanding.
Overall, MMASD can be used to help therapists and researchers to understand children's cognitive status, track development progress in therapy, and guide the treatment plan accordingly. 
It also provides inspiration for downstream tasks, such as activity recognition \cite{Pandey2020Guided}, action quality assessment \cite{Li2021improving}, and interpersonal synchrony estimation \cite{li2022pose}. 
In contrast to current datasets, e.g., listed in Table ~\ref{tab:dataset}, our dataset stands out for the following reasons:
\begin{itemize}
    \item It is a \textbf{publicly accessible} benchmark dataset for movement and behavior analysis during therapeutic interventions featured by diverse scenes, group activities, and participants.
    \item It includes \textbf{multimodal} features such as optical flow, 2D/3D skeletons, demographic, and clinical evaluation data. These features provide \textbf{privacy-preserving} approaches to maintain critical and full-body motion information. 
    \item Each scene depicts the same activity performed by a child and one or more therapists, providing an in-place template for comparing typically developing individuals with children with ASD. 
\end{itemize}

The rest of the paper is organized as follows. 
Further background on relevant autism datasets is presented in Section~\ref{sec:relatedwork}, followed by a presentation of data collection approach in Section~\ref{sec:method}. 
Details of the dataset, including statistics, data processing and annotation, are provided in Section~\ref{sec:dataset}. Finally, discussion and current limitations of MMASD are described in Section~\ref{sec:discussion} and conclusion in Section~\ref{sec:conclusion}.

\begin{table*}
  \caption{A comparison of related benchmarks and our dataset focused on behavior and movement analysis of children with ASD, including the target population, research focus, data modalities, computational models, data sample, and availability.}
  \label{tab:dataset}
  \resizebox{\linewidth}{!}{
  \begin{tabular}{lp{1.2cm}p{2.5cm}p{3.2cm}p{3.4cm}p{3.4cm}p{1.38cm}}
  \hline
    Dataset  & Target  \newline Population & Research Focus  & Data Modalities & Computational Models & $N$ of Data Samples \newline (Data length)* & Availability   \\  
    \hline
    Billing \emph{et al.} \cite{billing2020dream}  
    & 61\,ASD & Behavior analysis & Body motion, head pose, eye gaze & SVM, face detector \cite{viola2004robust}, Microsoft Kinect SDK & 3,121 sessions (306\,h) & Public \\
    
    Rajagopalan \emph{et al.} \cite{rajagopalan2013self} 
    & - 
    & Self-stimulatory behaviors detection & Video, audio & Space time interest point (STIP) \cite{laptev2005space} & 75 videos (90\,s/video) & Public\\
    
    Rehg \emph{et al.} \cite{rehg2013decoding} 
    & 121\,total & Behavior analysis & Video, audio, physiological data & SVM, Omron OKAO vision library & 160 sessions (3--5\,m/video) & On request \\

    DE-ENIGMA \cite{riva2020enigma}  & 128\,ASD & Behavior analysis & Facial landmark, body postures, video, audio & Computer vision, Microsoft Kinect SDK & 50 annotated recordings (154\,h) & On request \\

    Zunino \emph{et al.} \cite{zunino2018video} & 20\,ASD, 20\, TD** & Grasping actions & Hand \& arm trajectories & LSTM, CNN, Histogram of optical flow (HOG), VLAD \cite{jegou2011aggregating} 
    & 1,837 videos (83 frames/video)  & On request \\
    
    Pandey \emph{et al.} \cite{Pandey2020Guided} 
    & 37\,ASD & Behavior analysis & Video, 2D pose, optical flow & Guided weak supervision, Temporal segment networks, Inflated 3D CNN & 1,481 video clips & On request \\

    Del Coco \emph{et al.} \cite{del2017study} & 8\,ASD & Behavior analysis  & Facial landmark, gaze, head pose & Constrained Local Neural Fields \cite{baltrusaitis2013constrained} 
    & 6 videos (joint activities), 4 videos (imitative play) & Private 
    \\

    Dawson \emph{et al.} \cite{dawson2018atypical} & 22\,ASD, 82\,TD & Phenotyping, head movement  & Head pose, facial landmark & Model-based object pose, Computer vision analysis (CVA) \cite{la2015intraface}  
    & 10 videos (1--3\,s/video, 49 facial landmarks)  & Private \\

    Martin \emph{et al.} \cite{martin2018objective} & 21\,ASD, 21\,TD & Head movement  & Head pose, facial landmark & Computer-vision head tracking (Zface) \cite{jeni2015dense}  
    & 252 videos (6\,m/video)  & Private \\
    
    \textbf{MMASD} (Ours) & 32 ASD & Movement \& social behavior analysis & Optical flow, 2D pose, 3D pose, Clinical score & ROMP \cite{ROMP}, OpenPose \cite{cao2019openpose}, Lucas-Kanade \cite{lucas1981iterative} & 1,315 videos (7\,s\,\&\,186 frames/video) & Public  \\
    \hline
\end{tabular}
}
\footnotesize
\textit{*Certain works provided information on the duration of individual videos, whereas others presented the overall length of the dataset.\\}
\textit{**TD: Typically Developing.}
\end{table*}

\section{Related Work}
\label{sec:relatedwork}
ASD is characterized by atypical movement patterns, such as repetitive movements, clumsiness, and difficulties with coordination. A deeper understanding of these movement patterns and their association with ASD can aid therapists, clinicians, and researchers in developing more effective interventions and therapies.
To this end, several datasets have been developed for movement analysis of children with ASD. These datasets typically involve collecting motion capture or sensor data from children while they perform various activities or specific tasks. The collected data is then analyzed to identify differences and patterns in movement between children with ASD and typically developing children.
In this section, we present an overview of existing datasets that focus on movement and behavior analysis of children with ASD.

\paragraph{Movement analysis with humanoid robot interactions datasets}
Marinoiu \emph{et al.} \cite{marinoiu20183d} presented a dataset and system that use a humanoid robot to interact with children with ASD and monitor their body movements, facial expressions, and emotional states. 
The results show that children significantly improved their ability to recognize emotions, maintain eye contact, and respond appropriately to social cues. They identified several effective techniques for detecting and analyzing the children's emotional and behavioral responses, such as analyzing the frequency and duration of specific behaviors.
Billing \emph{et al.} \cite{billing2020dream} proposed a dataset of behavioral data recorded from 61 children with ASD during a large-scale evaluation of robot-enhanced therapy.
The dataset comprises sessions where children interacted with a robot under the guidance of a therapist and sessions where children interacted one-on-one with a therapist.
For each session, they used three RGB and two RGBD (Kinect) cameras to provide detailed information, including body motion, head position and orientation, and eye gaze of children's behavior during therapy. 

Another dataset related to humanoid robot interactions with ASD children is DE-ENIGMA \cite{riva2020enigma}, which includes using a multimodal human-robot interaction system to teach and expand social imagination among children with ASD. 
The DE-ENIGMA dataset comprises behavioral features such as facial mapping coordinates, visual and auditory, and facilitates communication and social interaction between the children and the robot.
The authors indicated that the DE-ENIGMA could be used as an effective tool for teaching and expanding social imagination in children with ASD. They also suggest that the usage of a multimodal human-robot interaction could be a promising approach for developing interventions for children with ASD that aim to improve their social skills and promote better social integration.

\paragraph{Eye movement and vocalization datasets}
Duan \emph{et al.} \cite{duan2019dataset} introduced a dataset of eye movement collected from children with ASD. The dataset includes 300 natural scene images and eye movement data from 14 children with ASD and 14 healthy individuals. It was created to facilitate research on the relationship between eye movements and ASD, with the goal of designing specialized visual attention models.
Baird \emph{et al.} \cite{baird2017automatic} introduced a dataset of vocalization recordings from children with ASD.
They also evaluated classification approaches from the spectrogram of autistic speech instances. Their results suggest that automatic classification systems could be used as a tool for aiding in the diagnosis and monitoring of ASD in children. 

\paragraph{Behavior analysis datasets}
For the action recognition dataset of children with ASD, Pandey \emph{et al.} \cite{Pandey2020Guided} proposed a dataset of video recording actions and a technique to automate the response of video recording scenes for human action recognition. 
They evaluated their technique on two skill assessments with autism datasets and a real-world dataset of 37 children with ASD. 
Rehg \emph{et al.} \cite{rehg2013decoding} introduced a publicly available dataset including over 160 sessions of child-adult interactions. 
They discussed the use of computer vision and machine learning techniques to analyze and understand children's social behavior in different contexts. 
They also identified technical challenges in analyzing various social behaviors, such as eye contact, smiling, and discrete behaviors. 
Rajagopalan \emph{et al.} \cite{rajagopalan2013self} explored the use of computer vision techniques to identify self-stimulatory behaviors in children with ASD. They also presented a self-stimulatory behavior dataset (SSBD) to assess the behaviors from video records of children with ASD in uncontrolled natural settings.
Their dataset comprised 75 videos grouped into three categories: arm flapping, head banging, and spinning behaviors. 

\paragraph{Comparison to \textbf{MMASD}}
In \autoref{tab:dataset}, we compare our proposed dataset with related benchmarks. 
Overall, MMASD features diverse themes and scenes, capturing full-body movements with multimodal features. 
In contrast, some works focused specifically on upper-body movements \cite{zunino2018video,del2017study,dawson2018atypical, martin2018objective}.  
MMASD also provides critical privacy-preserving features to represent body movements making it publicly accessible, while some works were conducted on raw videos that are either private or accessible only upon request  \cite{billing2020dream, rehg2013decoding, riva2020enigma, del2017study,dawson2018atypical, martin2018objective}.
Additionally, it is collected from therapeutic interventions, reflecting participants' motor ability and providing valuable insights for treatment guidance.

\section{Method}
\label{sec:method}
In the following sections, we describe the participants, procedure, and experimental settings of our proposed dataset. This study was approved by the University of Delaware's Institutional Review Board (IRB) \# $637082-12$.

\subsection{Participants}
We recruited $32$ children with ASD (27 males and 5 females)  from different races (Caucasian, African American, Asian, and Hispanic) 
through flyers posted online and onsite in local schools, services, and self-advocacy groups. 
Prior to enrollment, children were screened using the Social Communication Questionnaire \cite{srinivasan2016effects}, and their eligibility was determined by the Autism Diagnostic Observation Schedule-2 (ADOS-2) \cite{lord2000autism, srinivasan2015comparison} as well as clinical judgment.
All the children were between 5 and 12 years old. 
Written parental consent was obtained before enrollment. 
The Vineland Adaptive Behavior Scales \cite{sparrow1989vineland} were used to assess the children's adaptive functioning levels. 
In general, 82\% of the participating children had delays in the Adaptive Behavior Composite. Specifically, 70\% of them experienced communication delays, 80\% had difficulties with daily living skills, and 82\% had delays in socialization 

\subsection{Procedure}
The study was conducted over ten weeks, with the pre-test and post-test being conducted during the first and last weeks of the study, respectively.
Each training session was scheduled four times per week and lasted approximately 45 minutes. 
During the intervention, the trainer and adult model interacted with the child within a triadic context, with the adult model acting as the child's confederate and participating in all activities with the child.
This triadic setting (child, trainer, and model) provided numerous opportunities for promoting social and fine motor skills such as eye contact, body gesturing and balancing, coordination, and interpersonal synchrony during joint action games. 

All expert trainers and models involved were either physical therapists or physical therapy/kinesiology graduate students who had received significant pediatric training prior to their participation. 
The trainers and models were unknown to the children before the study. 
In addition to the expert training sessions, we also encouraged parents to provide two additional weekly sessions involving similar activities to promote practice. 
Parents were provided with essential instruction manuals, supplies, and in-person training beforehand. 
All training sessions were videotaped with the parents' consent and notification to the children, and the training diary was compiled by parents in collaboration with expert trainers. 
The general pipeline of training sessions had a standard procedure despite some unique activities across different themes. 
A welcoming and debriefing phase was present at the beginning and end of the data collection to help children warm up and get ready for the intervention, as well as to facilitate the subsequent data processing stage by providing time labels that indicate the segments to investigate. 

\subsection{Experiment Settings}
\begin{figure*}
    \centering
    \includegraphics[width=\linewidth]{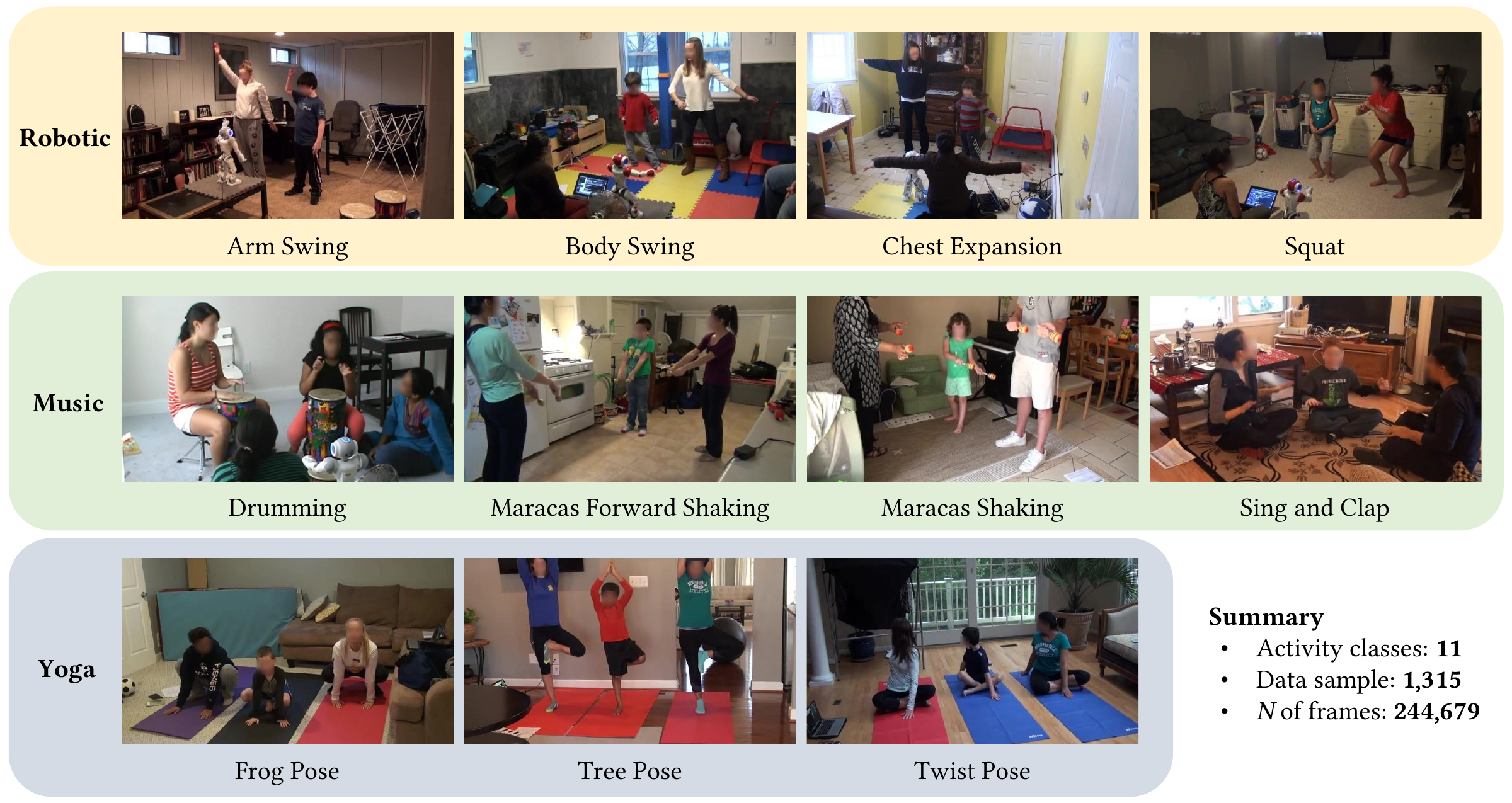}
    \caption{Sample scenes depicting various themes and activity classes present in the MMASD dataset.}
    \label{fig:dataset}
\end{figure*}

All videos from triadic settings were recorded in a house environment with the camera pointed toward the participating child.
Different tools were introduced to facilitate the training process depending on the theme of the intervention, for example, instruments and robots. 
Selected scenes in different themes of our proposed MMASD dataset are shown in \autoref{fig:dataset}.

\section{MMASD Dataset}
\label{sec:dataset}
MMASD includes 32 children diagnosed with autism of different levels.
It covers three unique themes:
\begin{itemize}
    \item Robot: children followed a robot and imitated bodily movements.
    \item Rhythm: children and therapists played musical instruments or sang together as a form of therapy.
    \item Yoga: children participated in yoga exercises led by therapists. These exercises included body stretching, twisting, balancing, and other activities. 
\end{itemize}


Overall, MMASD comprises 1,315 video clips that have been meticulously gathered from intervention video recordings spanning more than 108 hours.
It consists of 244,679 frames with an average duration of 7.15 seconds. 
The average data length in MMASD is $7.0 \pm 3.4$ seconds ($186.1 \pm 92.9$ frames), with dimensions ranging from $320\times240$ to $1920\times1080$.
\autoref{tab:Statistics} presents statistical information on MMASD. 
Depending on the conducted activity during the intervention, we further categorized all data into eleven activity classes as described in \autoref{tab:class description}. 
Each activity class falls into a unique theme, as shown in \autoref{fig:dataset}.
MMASD also reports demographic and autism evaluation scores of all participating children, including date of birth, motor functioning score, and severity of autism.


\begin{table}[!t]
    \caption{Statistics of our proposed MMASD dataset.}
    \centering
    \resizebox{\linewidth}{!}{
    \begin{tabular}{ll}
        \hline
         Description & Value \\
        \hline
       Number of data samples & 1,315 \\
        Number of frames & 244,679\\ 
         Number of activity classes & 11\\
        Average video length (seconds) &  $7.0 \pm 3.4$\\
        Average number of frames & $186.1 \pm 92.9$ \\
         Resolution &  $320\times240$ $\sim$ $1920\times1080$\\
         FPS & 25 $\sim$ 30\\
    \hline
    \end{tabular}}
    \label{tab:Statistics}
\end{table}

\begin{table*}
  \caption{Description and distribution of all 11 activity classes in MMASD.}
  \label{tab:class description}
  \begin{tabular}{{p{3.2cm}p{12.5cm}p{1cm}}}
  \hline
    Activity Class & Activity Description & Count  \\  
    \hline
    \emph{Arm swing} & The participant raises their left and right arm in succession while maintaining an upright posture. & 105  \\ 
    \emph{Body swing} & The participant swings their body left and right while stretching out both hands, one behind the other. & 119\\
    \emph{Chest expansion}& The participant gradually opens and closes their chest. & 114\\
    \emph{Drumming} & The participant plays either the snare or Tubano drum with one or both hands. & 168\\
    \emph{Frog pose} & The participant widens their knees as far as possible, places their feet with the big toes touching their body, and assumes a shape like that of a frog in the kneeling position. & 113\\   
    \emph{Maracas forward shaking} & The participant shakes maracas back and forth, an instrument commonly appearing in Caribbean and Latin music. & 103\\
    \emph{Maracas shaking} & The participant shakes maracas left and right in front of their chest. & 130\\
   \emph{Sing and clap} & The participant sits on the ground while simultaneously singing and clapping, typically done at the start or end of an intervention.  & 113\\
    \emph{Squat} & The participant repeatedly performs a crouching stance with their knees bent. & 101\\
    \emph{Tree pose} & The participant balances on one leg and places the sole of the other foot on the inner thigh, calf, or ankle of the standing leg in tree pose. & 129\\
    \emph{Twist pose} & The participant sits with their legs crossed and twists their torso to one side, keeping their lower body stable and grounded. & 120\\
    \hline
\end{tabular}
\end{table*}

\subsection{Data processing}
From original video recordings, we manually find out the start and end time stamps of a specific activity. Then we segmented the video into clips and categorized them by activity class.
Clips shorter than three seconds were discarded. We also discarded noisy data due to video quality, lighting conditions, and body occlusion.
Besides all the eleven activities in MMASD, there were some other activities with fewer examples. 
To ensure a balanced data distribution, we 
excluded all inadequate classes.

\subsection{Data Annotation}
We have four annotators that are well-trained in intervention understanding. 
The annotators had a comprehensive understanding of interventions and a cross-disciplined background in computer science and physical therapy. 
We exclusively assigned one activity class label to each video.
Each annotator completed data annotation independently, and the final class label was determined by majority voting. 
However, original videos cannot be publicly shared due to privacy concerns.
Therefore, we create data samples from every video clip by extracting selected features from the original scenes, including (1) optical flow, (2) 2D skeleton, and (3) 3D skeleton, respectively. 
All these features can maintain critical body movements while preserving privacy. 
Section \ref{sec:feat_extract} explained all selected features in detail. 

In addition to the motion-related features mentioned above, we also reported clinician evaluation results such as motor functioning score, and severity of autism for each participating child.  
ADOS-2 is a standardized assessment tool used to evaluate individuals suspected of having ASD. 
It is used in conjunction with other diagnostic information to help clinicians determine whether an individual meets the criteria for an ASD diagnosis.
ADOS-2 includes several modules, each designed for individuals of different ages and language abilities and includes a series of activities and tasks that are used to observe key features of ASD, such as social communication skills and repetitive behaviors.
The ADOS-2 scores are based on an individual's performance during activities and tasks specific to their module and can range from 0 to 10 or higher depending on the algorithm used, with higher scores indicating more severe ASD symptoms.
Moreover, we 
reported the ADOS comparison score, a continuous metric ranging from 1 to 10 that describes the severity of a child's autism symptoms compared to children with ASD of similar age and language levels \cite{gotham2009standardizing}. 
Low comparison scores are indicative of minimal evidence of autism symptoms, whereas high scores are indicative of severe autism symptoms.

The contrast between the ADOS-2 score and the ADOS comparison score is worth mentioning. 
The ADOS-2 score reflects an individual's raw score on the ADOS-2 assessment tool, while the ADOS comparison score is a statistical measure that compares an individual's performance to others of the same age and language level. 
The motor functioning score refers to an assessment of an individual's motor skills and abilities and is evaluated based on the children's level of independence in daily living skills \cite{srinivasan2015effects,srinivasan2016effects}.
It is on a scale of 1 to 3, while $1, 2, 3 $ represent low functioning (needing significant support), medium functioning (needing moderate support), and high functioning (needing less support), respectively. 
Finally, the severity of autism is determined by a comprehensive assessment that includes both the ADOS and motor function evaluation.

\subsection{Multimodal Feature Extraction}
\label{sec:feat_extract}
In order to preserve critical details of movement while avoiding any infringement of privacy, we derived the subsequent features from the initial footage.

\textbf{(1) Optical flow}
An optical flow is commonly referred to as the apparent motion of individual pixels between two consecutive frames on the image plane. 
Optical flow derived from raw videos (see Figure \ref{fig:teaser}) can provide a concise description of both the region and velocity of a motion without exposing an individual's identity \cite{carreira2017i3d,Feichtenhofer2016twostream,gao2020asymmetric, Pandey2020Guided}. 

\textbf{(2) 2D Skeleton} 
Skeleton data has an edge over RGB representations because it solely comprises the 2D positions of the human joints, which offer highly conceptual and context-independent data. 
This allows models to concentrate on the resilient aspects of body movements.
2D skeleton data has been widely applied to tasks relating to human behavior understanding, such as action recognition \cite{zheng2021poseformer,yan2018spatial}, action quality assessment \cite{li2022pose,li2022dyadic} and beyond.
An optimal way to acquire skeleton data is through the use of wearable devices and sensors that are affixed to the human body.
However, in the context of autism research, it poses a substantial challenge as children may feel overwhelmed wearing these devices and experience anxiety.  
As a result, the skeleton data extraction process is carried out by pre-trained pose detectors based on deep neural networks.

\textbf{(3) 3D Skeleton}
Similar to 2D skeleton data, 3D skeletons instead represent each key joint with a 3D coordination, introducing an additional depth dimension.
Since all the data was collected using a single RGB camera, we also completed this process with the help of deep neural networks. 

The technical details and tools utilized for feature extraction can be found in Section \ref{sec:implementation}.

\subsection{Data Format}
Suppose the original video clip includes $N$ participants (child, trainer, and assistant) 
is composed of $L$ frames, and the height and width of each frame are $H$ and $W$, respectively. 
As discussed above, each \textit{data sample} consists of four distinct components, with data dimension demonstrated in braces:
\begin{itemize}
    \item Optical flow $(L-1, H, W)$: saved as \textit{npy} files \cite{harris2020numpy}.
    \item 2D skeleton $(L, N , 17, 2)$: 2D coordinates of 17 key joints, following COCO \cite{coco2014cross} format, saved as \textit{JSON} files.
    \item 3D skeleton $(L, N, 24, 3)$: 3D coordinates of 24 key joints, following ROMP \cite{ROMP} format, saved as \textit{npz} files.
    \item Demographic and clinical evaluation for ASD $(9,)$: including nine attributes such as participant ID, date of birth, chronological age, social affect score, restricted and repetitive behavior score, motor functioning score, and severity of autism, saved as \textit{CSV} files.
\end{itemize}

\subsection{Implementation Details}
\label{sec:implementation}
\subsubsection{Optical flow:} The Lucas-Kanade~\cite{lucas1981iterative} method is used for our study. 
It is a popular technique used in computer vision to estimate the motion of objects between consecutive frames. 
The method assumes that the displacement of the image contents between two nearby instants is small and approximately constant within a neighborhood of the point under consideration. By solving the optical flow equation for all pixels within a window centered at the point, the method can estimate the motion of objects in the image sequence. Overall, the Lucas-Kanade optical flow method is an effective and preferred technique for estimating motion in various computer vision applications.
\subsubsection{2D skeleton:} The OpenPose method \cite{cao2019openpose} is used to extract 2D skeletons from our dataset of human action videos. 
OpenPose is a powerful tool for body, face, and hand analysis, developed by Carnegie Mellon University, and is based on Convolutional Neural Networks (CNNs).
It is a real-time multi-person key-point detection library that can accurately detect the key points of a human body, including joints and body parts, from an image or video feed. 
Initially, it predicts confidence maps for every body part and subsequently associates them with distinct individuals via Part Affinity Fields.
The library is open-source and written in C++ with a Python API, which makes it easy to use and integrate into various computer vision applications.

\subsubsection{3D skeleton:} We utilized the Regression of Multiple 3D People (ROMP) proposed by Sun \emph{et al.} \cite{ROMP}, a state-of-the-art technique to estimate the depth and pose of an individual from a single 2D image. The authors proposed a deep learning-based approach that is based on a fully convolutional architecture, which takes an input image and directly predicts the 3D locations of the body joints of the person(s) present in the image. This is achieved by directly estimating multiple differentiable maps from the entire image, which includes a Body Center heatmap and a Mesh Parameter map. 3D body mesh parameter vectors of all individuals can be extracted from these maps using a simple parameter sampling process. These vectors are then fed into the SMPL body model to generate multi-person 3D meshes.

In our study, we employed the code and pre-trained model shared by the authors and used it on our dataset to suit our specific needs. By utilizing this method and applying it to our own data, we obtained 2D and 3D coordinates of key joints of the person(s). Since it is suitable for occluded scenes and noisy data, ROMP demonstrated its ability to successfully identify and represent the dynamics of our multi-class, multi-person activities. 

\section{Discussion}
\label{sec:discussion}
This section delves into the challenges, insights, and future opportunities of MMASD dataset.
As the experiments were conducted in real-world settings in children's homes, we faced common computer vision challenges, including varying video quality, illumination changes, cluttered backgrounds, and pose variations. 
Notably, in the feature extraction stage, we encountered pose detection failures in challenging scenarios, such as body occlusion and participants moving out of the scene.
The intrinsic video quality limitation of MMASD also restricted us from capturing subtle and fine-grained features, such as facial expressions.

Moreover, it is imperative to conduct in-depth investigations into domain-specific challenges. 
For instance, in standard benchmarks for human activity recognition, typically developing individuals exhibit dominant and continuous actions with similar intensity. However, data on MMASD may not solely contain the target behavior throughout the entire duration, as impromptu actions or distractions may (and will) occur during therapy sessions for children with ASD.
Furthermore, unlike prevailing benchmarks that collect ground truth skeleton data by attaching sensors to the human body, MMASD generates skeleton data by means of pre-trained deep neural networks.  
This is because children with autism have limited tolerance for external stimuli, and the presence of sensors on their bodies may cause them to become anxious, agitated, or exhibit challenging behaviors.
Consequently, the skeleton data's reliability in MMASD depends on the performance of the underlying pose detectors. In addition, children with autism can exhibit varying motor functions, resulting in different intensity levels and completion rates for the same activity. 

There are several directions for future work that are worth exploring.
Firstly, further research can be conducted to develop and compare machine learning models on the MMASD dataset for various tasks, such as action quality assessment \cite{Li2021improving}, interpersonal synchrony estimation \cite{li2022dyadic,li2022pose}, and cognitive status tracking \cite{li2023facealignment, chheang2023towards}. 
This can help establish benchmark performance and identify state-of-the-art methods for analyzing the full-body movements of children  during therapeutic interventions.
In addition, new approaches can be investigated to overcome pose detection failures in MMASD. For example, by introducing pose uncertainty \cite{li2022pose} or attention mechanism to assign higher weights to more reliable body joints.
Furthermore, the MMASD dataset can be expanded in several aspects. 
This includes new features such as mutual gaze \cite{guo2023social}, or additional annotations including movement synchrony scores or task-specific clinician evaluations. Also, the MMASD dataset can be expanded by the inclusion of (age- and gender-matched) typically developing children data for developing more comprehensive and contrastive models.
Finally, efforts can be made to dataset augmentation via existing benchmarks not limited to the autism domain by matching samples with similar motion features \cite{Pandey2020Guided}, which can significantly expand the scale of the autism dataset.

\section{Conclusion}
\label{sec:conclusion}
Autism research has been greatly facilitated by machine learning techniques, which offer cost-effective, non-invasive, and accurate ways to analyze various aspects of children's behavior and development. 
However, the lack of open-access datasets has posed challenges to conducting fair comparisons and promoting sound research practices in the field of autism research. 
In this paper, we have proposed the MMASD, a privacy-preserving publicly accessible multimodal ASD children dataset $(N=32)$. The dataset features diverse, hand-annotated clips from over $100 hrs$ of raw videos from the play therapy interventions.
Our dataset includes multimodal features such as 2D \& 3D skeleton data, optical flow, demographic data, and clinical rating, offering a confidential data-sharing approach that can maintain critical full-body motion information.
Moreover, each scene in our dataset depicts the same activity performed by a child and one or more therapists, providing a valuable template for comparing typically developing individuals with children with ASD. 
The open-access MMASD dataset distinguishes itself from existing works by utilizing privacy-preserving multimodal features to provide comprehensive representations of full-body movements across diverse therapeutic social activities.

\begin{acks}
We would like to express our sincere gratitude for our research
team, therapists
and students who conducted the therapy sessions (Sudha Srinivasan, Maninderjit Kaur, and Isabel Park). We also thank therapy participants and their caregivers for contributing to the MMASD dataset
collection. This work is partly supported by our sponsors, Amazon
Research Awards Program, University of Delaware AI Center of Excellence, the National Institute of General Medical Sciences (NIGMS, P20 GM103446E), the National Institutes of Mental Health
(NIMH, 5R21MH089441-02, 4R33MH089441-03), and Autism Speaks
(Grant \#8137). Any opinions, findings, conclusions, or recommendations expressed in this paper are those of the authors and do not
necessarily reflect the views of the sponsors.
\end{acks}

\balance
\bibliographystyle{ACM-Reference-Format}
\bibliography{ref}


\begin{thebibliography}{45}


\ifx \showCODEN    \undefined \def \showCODEN     #1{\unskip}     \fi
\ifx \showDOI      \undefined \def \showDOI       #1{#1}\fi
\ifx \showISBNx    \undefined \def \showISBNx     #1{\unskip}     \fi
\ifx \showISBNxiii \undefined \def \showISBNxiii  #1{\unskip}     \fi
\ifx \showISSN     \undefined \def \showISSN      #1{\unskip}     \fi
\ifx \showLCCN     \undefined \def \showLCCN      #1{\unskip}     \fi
\ifx \shownote     \undefined \def \shownote      #1{#1}          \fi
\ifx \showarticletitle \undefined \def \showarticletitle #1{#1}   \fi
\ifx \showURL      \undefined \def \showURL       {\relax}        \fi
\providecommand\bibfield[2]{#2}
\providecommand\bibinfo[2]{#2}
\providecommand\natexlab[1]{#1}
\providecommand\showeprint[2][]{arXiv:#2}

\bibitem[Baio et~al\mbox{.}(2018)]%
        {baio2018prevalence}
\bibfield{author}{\bibinfo{person}{Jon Baio}, \bibinfo{person}{Lisa Wiggins}, \bibinfo{person}{Deborah~L Christensen}, \bibinfo{person}{Matthew~J Maenner}, \bibinfo{person}{Julie Daniels}, \bibinfo{person}{Zachary Warren}, \bibinfo{person}{Margaret Kurzius-Spencer}, \bibinfo{person}{Walter Zahorodny}, \bibinfo{person}{Cordelia~Robinson Rosenberg}, \bibinfo{person}{Tiffany White}, {et~al\mbox{.}}} \bibinfo{year}{2018}\natexlab{}.
\newblock \showarticletitle{Prevalence of autism spectrum disorder among children aged 8 years—autism and developmental disabilities monitoring network, 11 sites, United States, 2014}.
\newblock \bibinfo{journal}{\emph{MMWR Surveillance Summaries}} \bibinfo{volume}{67}, \bibinfo{number}{6} (\bibinfo{year}{2018}), \bibinfo{pages}{1}.
\newblock


\bibitem[Baird et~al\mbox{.}(2017)]%
        {baird2017automatic}
\bibfield{author}{\bibinfo{person}{Alice Baird}, \bibinfo{person}{Shahin Amiriparian}, \bibinfo{person}{Nicholas Cummins}, \bibinfo{person}{Alyssa~M Alcorn}, \bibinfo{person}{Anton Batliner}, \bibinfo{person}{Sergey Pugachevskiy}, \bibinfo{person}{Michael Freitag}, \bibinfo{person}{Maurice Gerczuk}, {and} \bibinfo{person}{Bj{\"o}rn Schuller}.} \bibinfo{year}{2017}\natexlab{}.
\newblock \showarticletitle{Automatic classification of autistic child vocalisations: A novel database and results}.
\newblock  (\bibinfo{year}{2017}).
\newblock


\bibitem[Baltrusaitis et~al\mbox{.}(2013)]%
        {baltrusaitis2013constrained}
\bibfield{author}{\bibinfo{person}{Tadas Baltrusaitis}, \bibinfo{person}{Peter Robinson}, {and} \bibinfo{person}{Louis-Philippe Morency}.} \bibinfo{year}{2013}\natexlab{}.
\newblock \showarticletitle{Constrained local neural fields for robust facial landmark detection in the wild}. In \bibinfo{booktitle}{\emph{Proceedings of the IEEE international conference on computer vision workshops}}. \bibinfo{pages}{354--361}.
\newblock


\bibitem[Billing et~al\mbox{.}(2020)]%
        {billing2020dream}
\bibfield{author}{\bibinfo{person}{Erik Billing}, \bibinfo{person}{Tony Belpaeme}, \bibinfo{person}{Haibin Cai}, \bibinfo{person}{Hoang-Long Cao}, \bibinfo{person}{Anamaria Ciocan}, \bibinfo{person}{Cristina Costescu}, \bibinfo{person}{Daniel David}, \bibinfo{person}{Robert Homewood}, \bibinfo{person}{Daniel Hernandez~Garcia}, \bibinfo{person}{Pablo G{\'o}mez~Esteban}, {et~al\mbox{.}}} \bibinfo{year}{2020}\natexlab{}.
\newblock \showarticletitle{The DREAM Dataset: Supporting a data-driven study of autism spectrum disorder and robot enhanced therapy}.
\newblock \bibinfo{journal}{\emph{PloS one}} \bibinfo{volume}{15}, \bibinfo{number}{8} (\bibinfo{year}{2020}), \bibinfo{pages}{e0236939}.
\newblock


\bibitem[Cao et~al\mbox{.}(2019)]%
        {cao2019openpose}
\bibfield{author}{\bibinfo{person}{Zhe Cao}, \bibinfo{person}{Gines Hidalgo}, \bibinfo{person}{Tomas Simon}, \bibinfo{person}{Shih-En Wei}, {and} \bibinfo{person}{Yaser Sheikh}.} \bibinfo{year}{2019}\natexlab{}.
\newblock \showarticletitle{OpenPose: realtime multi-person 2D pose estimation using Part Affinity Fields}.
\newblock \bibinfo{journal}{\emph{IEEE transactions on pattern analysis and machine intelligence}} \bibinfo{volume}{43}, \bibinfo{number}{1} (\bibinfo{year}{2019}), \bibinfo{pages}{172--186}.
\newblock


\bibitem[Carreira and Zisserman(2017)]%
        {carreira2017i3d}
\bibfield{author}{\bibinfo{person}{Joao Carreira} {and} \bibinfo{person}{Andrew Zisserman}.} \bibinfo{year}{2017}\natexlab{}.
\newblock \showarticletitle{Quo vadis, action recognition? a new model and the kinetics dataset}. In \bibinfo{booktitle}{\emph{proceedings of the IEEE Conference on Computer Vision and Pattern Recognition}}. \bibinfo{pages}{6299--6308}.
\newblock


\bibitem[Chen and Zhao(2019)]%
        {Chen2019AttentionBasd}
\bibfield{author}{\bibinfo{person}{Shi Chen} {and} \bibinfo{person}{Qi Zhao}.} \bibinfo{year}{2019}\natexlab{}.
\newblock \showarticletitle{Attention-Based Autism Spectrum Disorder Screening With Privileged Modality}. In \bibinfo{booktitle}{\emph{Proceedings of the IEEE/CVF International Conference on Computer Vision (ICCV)}}.
\newblock


\bibitem[Chheang et~al\mbox{.}(2023)]%
        {chheang2023towards}
\bibfield{author}{\bibinfo{person}{Vuthea Chheang}, \bibinfo{person}{Rommy Marquez-Hernandez}, \bibinfo{person}{Megha Patel}, \bibinfo{person}{Danush Rajasekaran}, \bibinfo{person}{Shayla Sharmin}, \bibinfo{person}{Gavin Caulfield}, \bibinfo{person}{Behdokht Kiafar}, \bibinfo{person}{Jicheng Li}, {and} \bibinfo{person}{Roghayeh~Leila Barmaki}.} \bibinfo{year}{2023}\natexlab{}.
\newblock \showarticletitle{Towards Anatomy Education with Generative AI-based Virtual Assistants in Immersive Virtual Reality Environments}.
\newblock \bibinfo{journal}{\emph{arXiv preprint arXiv:2306.17278}} (\bibinfo{year}{2023}).
\newblock


\bibitem[Coco and Dale(2014)]%
        {coco2014cross}
\bibfield{author}{\bibinfo{person}{Moreno~I Coco} {and} \bibinfo{person}{Rick Dale}.} \bibinfo{year}{2014}\natexlab{}.
\newblock \showarticletitle{Cross-recurrence quantification analysis of categorical and continuous time series: an R package}.
\newblock \bibinfo{journal}{\emph{Frontiers in psychology}}  \bibinfo{volume}{5} (\bibinfo{year}{2014}), \bibinfo{pages}{510}.
\newblock


\bibitem[Dawson et~al\mbox{.}(2018)]%
        {dawson2018atypical}
\bibfield{author}{\bibinfo{person}{Geraldine Dawson}, \bibinfo{person}{Kathleen Campbell}, \bibinfo{person}{Jordan Hashemi}, \bibinfo{person}{Steven~J Lippmann}, \bibinfo{person}{Valerie Smith}, \bibinfo{person}{Kimberly Carpenter}, \bibinfo{person}{Helen Egger}, \bibinfo{person}{Steven Espinosa}, \bibinfo{person}{Saritha Vermeer}, \bibinfo{person}{Jeffrey Baker}, {et~al\mbox{.}}} \bibinfo{year}{2018}\natexlab{}.
\newblock \showarticletitle{Atypical postural control can be detected via computer vision analysis in toddlers with autism spectrum disorder}.
\newblock \bibinfo{journal}{\emph{Scientific reports}} \bibinfo{volume}{8}, \bibinfo{number}{1} (\bibinfo{year}{2018}), \bibinfo{pages}{17008}.
\newblock


\bibitem[de~Belen et~al\mbox{.}(2020)]%
        {de2020computer}
\bibfield{author}{\bibinfo{person}{Ryan Anthony~J de Belen}, \bibinfo{person}{Tomasz Bednarz}, \bibinfo{person}{Arcot Sowmya}, {and} \bibinfo{person}{Dennis Del~Favero}.} \bibinfo{year}{2020}\natexlab{}.
\newblock \showarticletitle{Computer vision in autism spectrum disorder research: a systematic review of published studies from 2009 to 2019}.
\newblock \bibinfo{journal}{\emph{Translational psychiatry}} \bibinfo{volume}{10}, \bibinfo{number}{1} (\bibinfo{year}{2020}), \bibinfo{pages}{333}.
\newblock


\bibitem[Del~Coco et~al\mbox{.}(2017)]%
        {del2017study}
\bibfield{author}{\bibinfo{person}{Marco Del~Coco}, \bibinfo{person}{Marco Leo}, \bibinfo{person}{Pierluigi Carcagn{\`\i}}, \bibinfo{person}{Francesca Fama}, \bibinfo{person}{Letteria Spadaro}, \bibinfo{person}{Liliana Ruta}, \bibinfo{person}{Giovanni Pioggia}, {and} \bibinfo{person}{Cosimo Distante}.} \bibinfo{year}{2017}\natexlab{}.
\newblock \showarticletitle{Study of mechanisms of social interaction stimulation in autism spectrum disorder by assisted humanoid robot}.
\newblock \bibinfo{journal}{\emph{IEEE Transactions on Cognitive and Developmental Systems}} \bibinfo{volume}{10}, \bibinfo{number}{4} (\bibinfo{year}{2017}), \bibinfo{pages}{993--1004}.
\newblock


\bibitem[Duan et~al\mbox{.}(2019)]%
        {duan2019dataset}
\bibfield{author}{\bibinfo{person}{Huiyu Duan}, \bibinfo{person}{Guangtao Zhai}, \bibinfo{person}{Xiongkuo Min}, \bibinfo{person}{Zhaohui Che}, \bibinfo{person}{Yi Fang}, \bibinfo{person}{Xiaokang Yang}, \bibinfo{person}{Jes{\'u}s Guti{\'e}rrez}, {and} \bibinfo{person}{Patrick~Le Callet}.} \bibinfo{year}{2019}\natexlab{}.
\newblock \showarticletitle{A dataset of eye movements for the children with autism spectrum disorder}. In \bibinfo{booktitle}{\emph{Proceedings of the 10th ACM Multimedia Systems Conference}}. \bibinfo{pages}{255--260}.
\newblock


\bibitem[Feichtenhofer et~al\mbox{.}(2016)]%
        {Feichtenhofer2016twostream}
\bibfield{author}{\bibinfo{person}{Christoph Feichtenhofer}, \bibinfo{person}{Axel Pinz}, {and} \bibinfo{person}{Andrew Zisserman}.} \bibinfo{year}{2016}\natexlab{}.
\newblock \showarticletitle{Convolutional Two-Stream Network Fusion for Video Action Recognition}. In \bibinfo{booktitle}{\emph{2016 IEEE Conference on Computer Vision and Pattern Recognition (CVPR)}}. \bibinfo{pages}{1933--1941}.
\newblock
\urldef\tempurl%
\url{https://doi.org/10.1109/CVPR.2016.213}
\showDOI{\tempurl}


\bibitem[Gao et~al\mbox{.}(2020)]%
        {gao2020asymmetric}
\bibfield{author}{\bibinfo{person}{Jibin Gao}, \bibinfo{person}{Wei-Shi Zheng}, \bibinfo{person}{Jia-Hui Pan}, \bibinfo{person}{Chengying Gao}, \bibinfo{person}{Yaowei Wang}, \bibinfo{person}{Wei Zeng}, {and} \bibinfo{person}{Jianhuang Lai}.} \bibinfo{year}{2020}\natexlab{}.
\newblock \showarticletitle{An asymmetric modeling for action assessment}. In \bibinfo{booktitle}{\emph{European Conference on Computer Vision}}. Springer, \bibinfo{pages}{222--238}.
\newblock


\bibitem[Gotham et~al\mbox{.}(2009)]%
        {gotham2009standardizing}
\bibfield{author}{\bibinfo{person}{Katherine Gotham}, \bibinfo{person}{Andrew Pickles}, {and} \bibinfo{person}{Catherine Lord}.} \bibinfo{year}{2009}\natexlab{}.
\newblock \showarticletitle{Standardizing ADOS scores for a measure of severity in autism spectrum disorders}.
\newblock \bibinfo{journal}{\emph{Journal of autism and developmental disorders}}  \bibinfo{volume}{39} (\bibinfo{year}{2009}), \bibinfo{pages}{693--705}.
\newblock


\bibitem[Guo et~al\mbox{.}(2023)]%
        {guo2023social}
\bibfield{author}{\bibinfo{person}{Zhang Guo}, \bibinfo{person}{Vuthea Chheang}, \bibinfo{person}{Jicheng Li}, \bibinfo{person}{Kenneth~E Barner}, \bibinfo{person}{Anjana Bhat}, {and} \bibinfo{person}{Roghayeh Barmaki}.} \bibinfo{year}{2023}\natexlab{}.
\newblock \showarticletitle{Social Visual Behavior Analytics for Autism Therapy of Children Based on Automated Mutual Gaze Detection}. In \bibinfo{booktitle}{\emph{Proceedings of the International Conference on Cooperative and Human Aspects of Software Engineering}} (Orlando, Florida) \emph{(\bibinfo{series}{CHASE '23})}.
\newblock


\bibitem[Harris et~al\mbox{.}(2020)]%
        {harris2020numpy}
\bibfield{author}{\bibinfo{person}{Charles~R. Harris}, \bibinfo{person}{K.~Jarrod Millman}, \bibinfo{person}{St{\'{e}}fan~J. van~der Walt}, \bibinfo{person}{Ralf Gommers}, \bibinfo{person}{Pauli Virtanen}, \bibinfo{person}{David Cournapeau}, \bibinfo{person}{Eric Wieser}, \bibinfo{person}{Julian Taylor}, \bibinfo{person}{Sebastian Berg}, \bibinfo{person}{Nathaniel~J. Smith}, \bibinfo{person}{Robert Kern}, \bibinfo{person}{Matti Picus}, \bibinfo{person}{Stephan Hoyer}, \bibinfo{person}{Marten~H. van Kerkwijk}, \bibinfo{person}{Matthew Brett}, \bibinfo{person}{Allan Haldane}, \bibinfo{person}{Jaime~Fern{\'{a}}ndez del R{\'{i}}o}, \bibinfo{person}{Mark Wiebe}, \bibinfo{person}{Pearu Peterson}, \bibinfo{person}{Pierre G{\'{e}}rard-Marchant}, \bibinfo{person}{Kevin Sheppard}, \bibinfo{person}{Tyler Reddy}, \bibinfo{person}{Warren Weckesser}, \bibinfo{person}{Hameer Abbasi}, \bibinfo{person}{Christoph Gohlke}, {and} \bibinfo{person}{Travis~E. Oliphant}.} \bibinfo{year}{2020}\natexlab{}.
\newblock \showarticletitle{Array programming with {NumPy}}.
\newblock \bibinfo{journal}{\emph{Nature}} \bibinfo{volume}{585}, \bibinfo{number}{7825} (\bibinfo{date}{Sept.} \bibinfo{year}{2020}), \bibinfo{pages}{357--362}.
\newblock
\urldef\tempurl%
\url{https://doi.org/10.1038/s41586-020-2649-2}
\showDOI{\tempurl}


\bibitem[J{\'e}gou et~al\mbox{.}(2011)]%
        {jegou2011aggregating}
\bibfield{author}{\bibinfo{person}{Herv{\'e} J{\'e}gou}, \bibinfo{person}{Florent Perronnin}, \bibinfo{person}{Matthijs Douze}, \bibinfo{person}{Jorge S{\'a}nchez}, \bibinfo{person}{Patrick P{\'e}rez}, {and} \bibinfo{person}{Cordelia Schmid}.} \bibinfo{year}{2011}\natexlab{}.
\newblock \showarticletitle{Aggregating local image descriptors into compact codes}.
\newblock \bibinfo{journal}{\emph{IEEE transactions on pattern analysis and machine intelligence}} \bibinfo{volume}{34}, \bibinfo{number}{9} (\bibinfo{year}{2011}), \bibinfo{pages}{1704--1716}.
\newblock


\bibitem[Jeni et~al\mbox{.}(2015)]%
        {jeni2015dense}
\bibfield{author}{\bibinfo{person}{L{\'a}szl{\'o}~A Jeni}, \bibinfo{person}{Jeffrey~F Cohn}, {and} \bibinfo{person}{Takeo Kanade}.} \bibinfo{year}{2015}\natexlab{}.
\newblock \showarticletitle{Dense 3D face alignment from 2D videos in real-time}. In \bibinfo{booktitle}{\emph{2015 11th IEEE international conference and workshops on automatic face and gesture recognition (FG)}}, Vol.~\bibinfo{volume}{1}. IEEE, \bibinfo{pages}{1--8}.
\newblock


\bibitem[la~Torre et~al\mbox{.}(2015)]%
        {la2015intraface}
\bibfield{author}{\bibinfo{person}{F~De la Torre}, \bibinfo{person}{WS Chu}, \bibinfo{person}{X Xiong}, \bibinfo{person}{F Vicente}, \bibinfo{person}{X Ding}, {and} \bibinfo{person}{J Cohn}.} \bibinfo{year}{2015}\natexlab{}.
\newblock \showarticletitle{Intraface}. In \bibinfo{booktitle}{\emph{IEEE International Conference on Face and Gesture Recognition}}.
\newblock


\bibitem[Laptev(2005)]%
        {laptev2005space}
\bibfield{author}{\bibinfo{person}{Ivan Laptev}.} \bibinfo{year}{2005}\natexlab{}.
\newblock \showarticletitle{On space-time interest points}.
\newblock \bibinfo{journal}{\emph{International journal of computer vision}}  \bibinfo{volume}{64} (\bibinfo{year}{2005}), \bibinfo{pages}{107--123}.
\newblock


\bibitem[Li et~al\mbox{.}(2023)]%
        {li2023facealignment}
\bibfield{author}{\bibinfo{person}{Jicheng Li}, \bibinfo{person}{Roghayeh~Leila Barmaki}, \bibinfo{person}{Li Zhu}, \bibinfo{person}{Korosh Vatanparvar}, \bibinfo{person}{Migyeong Gwak}, \bibinfo{person}{Jilong Kuang}, {and} \bibinfo{person}{Alex Gao}.} \bibinfo{year}{2023}\natexlab{}.
\newblock \showarticletitle{Advancements in Face Alignment Evaluation for Contact-less Vital Sign Detection}. In \bibinfo{booktitle}{\emph{2023 IEEE-EMBS International Conference on Body Sensor Networks: Sensor and Systems for Digital Health (BSN)}}.
\newblock


\bibitem[Li et~al\mbox{.}(2021a)]%
        {Li2021improving}
\bibfield{author}{\bibinfo{person}{Jicheng Li}, \bibinfo{person}{Anjana Bhat}, {and} \bibinfo{person}{Roghayeh Barmaki}.} \bibinfo{year}{2021}\natexlab{a}.
\newblock \showarticletitle{Improving the Movement Synchrony Estimation with Action Quality Assessment in Children Play Therapy}. In \bibinfo{booktitle}{\emph{Proceedings of the International Conference on Multimodal Interaction}} (Montr\'{e}al, QC, Canada) \emph{(\bibinfo{series}{ICMI '21})}. \bibinfo{pages}{397–406}.
\newblock


\bibitem[Li et~al\mbox{.}(2021b)]%
        {li21twostage}
\bibfield{author}{\bibinfo{person}{Jicheng Li}, \bibinfo{person}{Anjana Bhat}, {and} \bibinfo{person}{Roghayeh Barmaki}.} \bibinfo{year}{2021}\natexlab{b}.
\newblock \showarticletitle{A Two-stage Multi-modal Affect Analysis Framework for Children with Autism Spectrum Disorder}. In \bibinfo{booktitle}{\emph{Proceedings of the AAAI-21 Workshop on Affective Content Analysis}} (New York, USA). \bibinfo{pages}{1--8}.
\newblock
\urldef\tempurl%
\url{http://ceur-ws.org/Vol-2897/AffconAAAI-21_paper1.pdf}
\showURL{%
\tempurl}


\bibitem[Li et~al\mbox{.}(2022a)]%
        {li2022dyadic}
\bibfield{author}{\bibinfo{person}{Jicheng Li}, \bibinfo{person}{Anjana Bhat}, {and} \bibinfo{person}{Roghayeh Barmaki}.} \bibinfo{year}{2022}\natexlab{a}.
\newblock \showarticletitle{Dyadic Movement Synchrony Estimation Under Privacy-preserving Conditions}. In \bibinfo{booktitle}{\emph{2022 26th International Conference on Pattern Recognition (ICPR)}}. IEEE, \bibinfo{pages}{762--769}.
\newblock


\bibitem[Li et~al\mbox{.}(2022b)]%
        {li2022pose}
\bibfield{author}{\bibinfo{person}{Jicheng Li}, \bibinfo{person}{Anjana Bhat}, {and} \bibinfo{person}{Roghayeh Barmaki}.} \bibinfo{year}{2022}\natexlab{b}.
\newblock \showarticletitle{Pose Uncertainty Aware Movement Synchrony Estimation via Spatial-Temporal Graph Transformer}. In \bibinfo{booktitle}{\emph{Proceedings of the International Conference on Multimodal Interaction}} (Bengaluru, India) \emph{(\bibinfo{series}{ICMI '22})}. \bibinfo{pages}{73–82}.
\newblock


\bibitem[Lord et~al\mbox{.}(2000)]%
        {lord2000autism}
\bibfield{author}{\bibinfo{person}{Catherine Lord}, \bibinfo{person}{Susan Risi}, \bibinfo{person}{Linda Lambrecht}, \bibinfo{person}{Edwin~H Cook}, \bibinfo{person}{Bennett~L Leventhal}, \bibinfo{person}{Pamela~C DiLavore}, \bibinfo{person}{Andrew Pickles}, {and} \bibinfo{person}{Michael Rutter}.} \bibinfo{year}{2000}\natexlab{}.
\newblock \showarticletitle{The Autism Diagnostic Observation Schedule—Generic: A standard measure of social and communication deficits associated with the spectrum of autism}.
\newblock \bibinfo{journal}{\emph{Journal of autism and developmental disorders}} \bibinfo{volume}{30}, \bibinfo{number}{3} (\bibinfo{year}{2000}), \bibinfo{pages}{205--223}.
\newblock


\bibitem[Lucas and Kanade(1981)]%
        {lucas1981iterative}
\bibfield{author}{\bibinfo{person}{Bruce~D Lucas} {and} \bibinfo{person}{Takeo Kanade}.} \bibinfo{year}{1981}\natexlab{}.
\newblock \showarticletitle{An iterative image registration technique with an application to stereo vision}. In \bibinfo{booktitle}{\emph{IJCAI'81: 7th international joint conference on Artificial intelligence}}, Vol.~\bibinfo{volume}{2}. \bibinfo{pages}{674--679}.
\newblock


\bibitem[Marinoiu et~al\mbox{.}(2018)]%
        {marinoiu20183d}
\bibfield{author}{\bibinfo{person}{Elisabeta Marinoiu}, \bibinfo{person}{Mihai Zanfir}, \bibinfo{person}{Vlad Olaru}, {and} \bibinfo{person}{Cristian Sminchisescu}.} \bibinfo{year}{2018}\natexlab{}.
\newblock \showarticletitle{3d human sensing, action and emotion recognition in robot assisted therapy of children with autism}. In \bibinfo{booktitle}{\emph{Proceedings of the IEEE conference on computer vision and pattern recognition}}. \bibinfo{pages}{2158--2167}.
\newblock


\bibitem[Martin et~al\mbox{.}(2018)]%
        {martin2018objective}
\bibfield{author}{\bibinfo{person}{Katherine~B Martin}, \bibinfo{person}{Zakia Hammal}, \bibinfo{person}{Gang Ren}, \bibinfo{person}{Jeffrey~F Cohn}, \bibinfo{person}{Justine Cassell}, \bibinfo{person}{Mitsunori Ogihara}, \bibinfo{person}{Jennifer~C Britton}, \bibinfo{person}{Anibal Gutierrez}, {and} \bibinfo{person}{Daniel~S Messinger}.} \bibinfo{year}{2018}\natexlab{}.
\newblock \showarticletitle{Objective measurement of head movement differences in children with and without autism spectrum disorder}.
\newblock \bibinfo{journal}{\emph{Molecular autism}}  \bibinfo{volume}{9} (\bibinfo{year}{2018}), \bibinfo{pages}{1--10}.
\newblock


\bibitem[Pandey et~al\mbox{.}(2020)]%
        {Pandey2020Guided}
\bibfield{author}{\bibinfo{person}{Prashant Pandey}, \bibinfo{person}{Prathosh AP}, \bibinfo{person}{Manu Kohli}, {and} \bibinfo{person}{Josh Pritchard}.} \bibinfo{year}{2020}\natexlab{}.
\newblock \showarticletitle{Guided Weak Supervision for Action Recognition with Scarce Data to Assess Skills of Children with Autism}.
\newblock \bibinfo{journal}{\emph{Proceedings of the AAAI Conference on Artificial Intelligence}} \bibinfo{volume}{34}, \bibinfo{number}{01} (\bibinfo{date}{Apr.} \bibinfo{year}{2020}), \bibinfo{pages}{463--470}.
\newblock
\urldef\tempurl%
\url{https://doi.org/10.1609/aaai.v34i01.5383}
\showDOI{\tempurl}


\bibitem[Rajagopalan et~al\mbox{.}(2013)]%
        {rajagopalan2013self}
\bibfield{author}{\bibinfo{person}{Shyam Rajagopalan}, \bibinfo{person}{Abhinav Dhall}, {and} \bibinfo{person}{Roland Goecke}.} \bibinfo{year}{2013}\natexlab{}.
\newblock \showarticletitle{Self-stimulatory behaviours in the wild for autism diagnosis}. In \bibinfo{booktitle}{\emph{Proceedings of the IEEE International Conference on Computer Vision Workshops}}. \bibinfo{pages}{755--761}.
\newblock


\bibitem[Rehg et~al\mbox{.}(2013)]%
        {rehg2013decoding}
\bibfield{author}{\bibinfo{person}{James Rehg}, \bibinfo{person}{Gregory Abowd}, \bibinfo{person}{Agata Rozga}, \bibinfo{person}{Mario Romero}, \bibinfo{person}{Mark Clements}, \bibinfo{person}{Stan Sclaroff}, \bibinfo{person}{Irfan Essa}, \bibinfo{person}{O Ousley}, \bibinfo{person}{Yin Li}, \bibinfo{person}{Chanho Kim}, {et~al\mbox{.}}} \bibinfo{year}{2013}\natexlab{}.
\newblock \showarticletitle{Decoding children's social behavior}. In \bibinfo{booktitle}{\emph{Proceedings of the IEEE conference on computer vision and pattern recognition}}. \bibinfo{pages}{3414--3421}.
\newblock


\bibitem[Riva et~al\mbox{.}(2020)]%
        {riva2020enigma}
\bibfield{author}{\bibinfo{person}{Giuseppe Riva}, \bibinfo{person}{Eleonora Riva}, {et~al\mbox{.}}} \bibinfo{year}{2020}\natexlab{}.
\newblock \showarticletitle{DE-ENIGMA: Multimodal Human-Robot Interaction for Teaching and Expanding Social Imagination in Autistic Children}.
\newblock \bibinfo{journal}{\emph{Cyberpsychology, behavior and social networking}} \bibinfo{volume}{23}, \bibinfo{number}{11} (\bibinfo{year}{2020}), \bibinfo{pages}{806--807}.
\newblock


\bibitem[Sparrow and Cicchetti(1989)]%
        {sparrow1989vineland}
\bibfield{author}{\bibinfo{person}{Sara~S Sparrow} {and} \bibinfo{person}{Domenic~V Cicchetti}.} \bibinfo{year}{1989}\natexlab{}.
\newblock \bibinfo{booktitle}{\emph{The Vineland adaptive behavior scales.}}
\newblock \bibinfo{publisher}{Allyn \& Bacon}.
\newblock


\bibitem[Srinivasan et~al\mbox{.}(2016)]%
        {srinivasan2016effects}
\bibfield{author}{\bibinfo{person}{Sudha~M Srinivasan}, \bibinfo{person}{Inge-Marie Eigsti}, \bibinfo{person}{Timothy Gifford}, {and} \bibinfo{person}{Anjana~N Bhat}.} \bibinfo{year}{2016}\natexlab{}.
\newblock \showarticletitle{The effects of embodied rhythm and robotic interventions on the spontaneous and responsive verbal communication skills of children with Autism Spectrum Disorder (ASD): A further outcome of a pilot randomized controlled trial}.
\newblock \bibinfo{journal}{\emph{Research in autism spectrum disorders}}  \bibinfo{volume}{27} (\bibinfo{year}{2016}), \bibinfo{pages}{73--87}.
\newblock


\bibitem[Srinivasan et~al\mbox{.}(2015a)]%
        {srinivasan2015effects}
\bibfield{author}{\bibinfo{person}{Sudha~M Srinivasan}, \bibinfo{person}{Maninderjit Kaur}, \bibinfo{person}{Isabel~K Park}, \bibinfo{person}{Timothy~D Gifford}, \bibinfo{person}{Kerry~L Marsh}, {and} \bibinfo{person}{Anjana~N Bhat}.} \bibinfo{year}{2015}\natexlab{a}.
\newblock \showarticletitle{The effects of rhythm and robotic interventions on the imitation/praxis, interpersonal synchrony, and motor performance of children with autism spectrum disorder (ASD): a pilot randomized controlled trial}.
\newblock \bibinfo{journal}{\emph{Autism research and treatment}}  \bibinfo{volume}{2015} (\bibinfo{year}{2015}).
\newblock


\bibitem[Srinivasan et~al\mbox{.}(2015b)]%
        {srinivasan2015comparison}
\bibfield{author}{\bibinfo{person}{Sudha~M Srinivasan}, \bibinfo{person}{Isabel~K Park}, \bibinfo{person}{Linda~B Neelly}, {and} \bibinfo{person}{Anjana~N Bhat}.} \bibinfo{year}{2015}\natexlab{b}.
\newblock \showarticletitle{A comparison of the effects of rhythm and robotic interventions on repetitive behaviors and affective states of children with Autism Spectrum Disorder (ASD)}.
\newblock \bibinfo{journal}{\emph{Research in autism spectrum disorders}}  \bibinfo{volume}{18} (\bibinfo{year}{2015}), \bibinfo{pages}{51--63}.
\newblock


\bibitem[Sun et~al\mbox{.}(2021)]%
        {ROMP}
\bibfield{author}{\bibinfo{person}{Yu Sun}, \bibinfo{person}{Qian Bao}, \bibinfo{person}{Wu Liu}, \bibinfo{person}{Yili Fu}, \bibinfo{person}{Black Michael~J.}, {and} \bibinfo{person}{Tao Mei}.} \bibinfo{year}{2021}\natexlab{}.
\newblock \showarticletitle{Monocular, One-stage, Regression of Multiple 3D People}. In \bibinfo{booktitle}{\emph{ICCV}}.
\newblock


\bibitem[Viola and Jones(2004)]%
        {viola2004robust}
\bibfield{author}{\bibinfo{person}{Paul Viola} {and} \bibinfo{person}{Michael~J Jones}.} \bibinfo{year}{2004}\natexlab{}.
\newblock \showarticletitle{Robust real-time face detection}.
\newblock \bibinfo{journal}{\emph{International journal of computer vision}}  \bibinfo{volume}{57} (\bibinfo{year}{2004}), \bibinfo{pages}{137--154}.
\newblock


\bibitem[Wall et~al\mbox{.}(2012)]%
        {wall2012use}
\bibfield{author}{\bibinfo{person}{Dennis~Paul Wall}, \bibinfo{person}{J Kosmicki}, \bibinfo{person}{TF Deluca}, \bibinfo{person}{E Harstad}, {and} \bibinfo{person}{Vincent~Alfred Fusaro}.} \bibinfo{year}{2012}\natexlab{}.
\newblock \showarticletitle{Use of machine learning to shorten observation-based screening and diagnosis of autism}.
\newblock \bibinfo{journal}{\emph{Translational psychiatry}} \bibinfo{volume}{2}, \bibinfo{number}{4} (\bibinfo{year}{2012}), \bibinfo{pages}{e100--e100}.
\newblock


\bibitem[Yan et~al\mbox{.}(2018)]%
        {yan2018spatial}
\bibfield{author}{\bibinfo{person}{Sijie Yan}, \bibinfo{person}{Yuanjun Xiong}, {and} \bibinfo{person}{Dahua Lin}.} \bibinfo{year}{2018}\natexlab{}.
\newblock \showarticletitle{Spatial temporal graph convolutional networks for skeleton-based action recognition}. In \bibinfo{booktitle}{\emph{Thirty-second AAAI conference on artificial intelligence}}.
\newblock


\bibitem[Zheng et~al\mbox{.}(2021)]%
        {zheng2021poseformer}
\bibfield{author}{\bibinfo{person}{Ce Zheng}, \bibinfo{person}{Sijie Zhu}, \bibinfo{person}{Matias Mendieta}, \bibinfo{person}{Taojiannan Yang}, \bibinfo{person}{Chen Chen}, {and} \bibinfo{person}{Zhengming Ding}.} \bibinfo{year}{2021}\natexlab{}.
\newblock \showarticletitle{3D Human Pose Estimation with Spatial and Temporal Transformers}.
\newblock \bibinfo{journal}{\emph{Proceedings of the IEEE International Conference on Computer Vision (ICCV)}} (\bibinfo{year}{2021}).
\newblock


\bibitem[Zunino et~al\mbox{.}(2018)]%
        {zunino2018video}
\bibfield{author}{\bibinfo{person}{Andrea Zunino}, \bibinfo{person}{Pietro Morerio}, \bibinfo{person}{Andrea Cavallo}, \bibinfo{person}{Caterina Ansuini}, \bibinfo{person}{Jessica Podda}, \bibinfo{person}{Francesca Battaglia}, \bibinfo{person}{Edvige Veneselli}, \bibinfo{person}{Cristina Becchio}, {and} \bibinfo{person}{Vittorio Murino}.} \bibinfo{year}{2018}\natexlab{}.
\newblock \showarticletitle{Video gesture analysis for autism spectrum disorder detection}. In \bibinfo{booktitle}{\emph{Proc. of International conference on pattern recognition (ICPR)}}. IEEE, \bibinfo{pages}{3421--3426}.
\newblock


\end{thebibliography}

\end{document}